\documentclass[sigconf,nonacm]{acmart}
\setkeys{acmart.cls}{balance=false}
\usepackage{booktabs}
\usepackage{microtype}
\usepackage{url}
\newcommand{\code}[1]{{\footnotesize\url{#1}}}

\title{Conditional Calibration Contracts under Coupled Load:
A Branch-Aware Audit of Switched Inference}
\author{Juan S. Santillana}
\email{juan@jsantillana.com}
\affiliation{\institution{Independent research}\city{}\country{}}

\begin{document}
\begin{abstract}
Bursty serving changes both queue states and the population selected for each
model. We study conditional calibration contracts as a specification to be
validated, rather than a certified property of an existing router. CA-MHD
combines load-coupled distillation, closed-loop replay and a cost-informed
Schmitt router in a discrete-event simulator with a hardware service table.
Across 40 cells (ten training seeds and four content-load coupling levels),
we distinguish the original teacher, frozen wide heads on an adapted trunk,
and the narrow student. A common teacher-trajectory audit has median absolute
student-teacher deficit 0.06 percentage points, but does not certify the
student-selected deployment support. Historical results use a kernel
correctness gap, not the published SmoothECE; we separately prove
selected-branch composition for a residual functional at a common fixed
scale. The deployed wide branch has lower ordinal accuracy than the original
teacher in all 40 cells. Same-checkpoint student-only serving has lower
historical ordinal gap in 27/40 cells; routing also exposes violation tails
up to 17.5 percent. Student p99 latency remains above the 1.5 ms target.
Retrospective temporal recalibration improves the ordinal gap in 39/40 cells,
but causal delayed-label availability is unverified and no incremental
per-mode benefit is detected after multiplicity adjustment. An in-sample
CUSUM detects 30/40 checkpoint changes at five historical false alarms,
with no prospective false-alarm guarantee. These findings support a
branch-aware audit and a scoped negative result, not demonstrated routing
superiority. Independent budgets, selected-support validation and net
utility under a satisfied SLA remain open.
\end{abstract}
\maketitle

\section{Introduction}
\label{sec:intro}
Elastic inference trades predictive performance against serving cost
\citep{clipper2017,infaas2021,chen2023frugalgpt}. Under bursty load,
the serving policy also changes queue history, batch sizes and which
events reach each predictor. A training-time calibration audit need not
describe that selected population. Moreover, a branch named ``teacher''
need not retain the original teacher when distillation updates a shared
trunk. These are specification and measurement problems before they are
arguments for switching.

We investigate conditional calibration contracts: versioned, per-cell
budgets for named predictors on a declared policy-induced support.
The existing campaign provides descriptive audits, not a validated
prospective contract. Calibration, accuracy, ordinal utility and latency
are separate obligations: equal accuracy does not imply equal probability
quality, and lower latency than the teacher does not imply SLA compliance.

\paragraph{Contributions and scope.}
\begin{enumerate}
\item A branch-aware specification separating the original teacher $T_0$,
the adapted wide branch $W$, and the student $S$, observed deficits from
independent budgets, and teacher-always from selected deployment support.
\item CA-MHD/CUL-Bench: coupled synthetic streams, load-coupled
distillation (LCD), DAgger-style replay and a cost-informed Schmitt router.
The 40-cell campaign extends an original three-seed study; historical
sweeps and ablations remain three-seed evidence.
\item Reconciled empirical diagnostics: common-trajectory comparisons,
fixed-state predictor counterfactuals, policy interventions, retrospective
recalibration and an in-sample monitoring curve. They reveal identity and
latency failure modes, not a universal advantage of routing.
\item Proved functional statements for a separate fixed-scale residual
metric, plus conditional switch-count and fluid-stability statements.
They do not certify the historical metric, the checkpoints or a p99 SLA.
\end{enumerate}

\section{Related Work}
\label{sec:related}
\paragraph{Elastic inference and serving.}
Slimmable networks \citep{yu2019slimmable,yu2019universally}, Once-for-All
\citep{cai2020ofa} and MatFormer \citep{devvrit2024matformer} support
budget-dependent subnetworks. SPINN \citep{laskaridis2020spinn} jointly
adapts early exits and device--cloud splitting to dynamic conditions and
service requirements; it is not merely a content-difficulty cascade.
Clipper, INFaaS and Clockwork
\citep{clipper2017,infaas2021,gujrati2020clockwork} address serving latency
and resource management. Our measured switching-cost ratio is a design
input, not a proof of optimal hysteresis or a unique systems ingredient.

\paragraph{Routing and contemporary load-aware scheduling.}
Hybrid LLM and RouteLLM \citep{ding2024hybrid,ong2024routellm} learn
quality/cost dispatchers. Selective classification and early exits
\citep{chow1970,geifman2017selective,teerapittayanon2016branchynet,
schuster2022confident} supply related input-level decision rules.
Workload-aware routing \citep{jain2024router} and NexusSched
\citep{zhang2025nexussched} explicitly use workload or state signals.
The recent measured routing study of Tumkur et al.
\citep{tumkur2026route} combines simulation with hardware validation and
reports regimes where routing benefits disappear or reverse. Its
``calibration'' concerns performance-time costs, not prediction
correctness. These antecedents preclude treating load awareness,
hardware measurement or regime sensitivity as novel in isolation.
We do not compare these systems under matched budgets, nor claim priority
from this non-exhaustive literature review.

\paragraph{Prediction calibration and monitoring.}
Dataset-shift robustness \citep{ovadia2019trust}, temperature scaling
\citep{guo2017calibration}, adaptive conformal inference
\citep{gibbs2021adaptive} and calibration-distance theory
\citep{blasiok2023unifying} provide distinct foundations.
SmoothECE \citep{blasiok2023smoothece} smooths residuals with a reflected
kernel and chooses its scale by a fixed point. Our historical correctness
gap differs and does not inherit its fidelity or sampling guarantees.
Farran's recent preprint \citep{farran2026monitor} studies anytime-valid
PIT-based monitoring; its announced guarantees cannot be transferred to
our overlapping-window, in-sample CUSUM without verifying assumptions.
The defensible contribution here is an integrated, branch-aware audit,
not a new Platt method, universal literature gap or established SOTA.

\section{Setup: Identity, Metrics and Contracts}
\label{sec:setup}
\paragraph{Serving model.}
Idle/burst arrival rates are 5,000/35,000 events/s, duty $D=0.25$,
period 8\,s and test horizon 16\,s. The hardware table gives approximately
495\,\textmu s for teacher and 431\,\textmu s for student at batch size 16;
$c_{\mathrm{sw}}=21.9$\,\textmu s is the nominal measured switch-cost input.
The DES samples lognormal service durations from this table, with forward
inference outside its service clock. Switch-cost variability is modeled;
the evicted-alternation microbench does not constitute a complete measured
distribution of live switch costs. The target is p99 $\le1.5$\,ms.
Full-batch burst loads $\rho_t=1.083$, $\rho_s=0.941$ describe a capacity
margin, not a proof of student SLA compliance or necessity of switching.
Teacher-only can be cycle-stable while accumulating large burst queues.

\paragraph{Predictor identity and eligibility.}
For each checkpoint version,
\[
 T_0=H_T\circ F_T,\qquad W=H_T\circ F_A,\qquad S=H_S\circ F_A.
\]
Student training updates the adapted trunk $F_A$ and narrow heads $H_S$,
while freezing wide heads $H_T$. Thus deployed teacher-mode serves $W$,
not $T_0$. LCD and LCD+DAgger are different versions. Identity includes
checkpoint/trunk/head hashes, preprocessing, temperature or confidence
map, PMF construction and tie rule. Serving audits use no temperatures;
training-protocol audits use their fitted temperatures.
Heads are binary gate, closed-set softmax choice, and ordinal corn
\citep{shi2023corn}. Corn is evaluated only on eligible attacks with valid
severity labels. Top-label correctness calibration does not certify the
whole ordinal PMF or its cumulative thresholds.

\paragraph{Historical kernel correctness gap.}
All historical quantities called \texttt{smoothECE}, \texttt{sECE} or
\texttt{sECE\_eval} in figures, tables and artifacts are \emph{legacy
identifiers} for the kernel correctness gap $H_\kappa$, not published
SmoothECE. For confidence $C$ and correctness $Y$, the pairwise definition
at radius $r=5$ is
\begin{align}
 K_{\kappa,r}(t,c)&=e^{-(t-c)^2/(2\kappa^2)}
                 \mathbf1\{|t-c|\le r\kappa\},\nonumber\\
 f_P(t)&=\frac{\mathbb E_P K_{\kappa,r}(t,C)Y}
                  {\mathbb E_P K_{\kappa,r}(t,C)},\qquad
 H_{\kappa,r}(P)=\mathbb E_P|C-f_P(C)|. \label{eq:historical}
\end{align}
We abbreviate this historical family as $H_\kappa$ ($\kappa=0.05$).
The implementation uses pairwise fitting with self-inclusion for
$n\le4096$, and a scatter/convolution/interpolated-correctness fit for
larger pools. The large-pool route ignores radius/self flags and has
displacement support approximately $[-0.5,0.5]$, not a general
$\pm5\kappa$ truncation. It is therefore a distinct numerical route,
not uniformly equivalent for all scales and sample sizes. These semantics
are retained rather than silently changing historical values.
For equal calibrated atoms at $C=0.5,0.6$, $H_{0.05}=0.0119203$ although
the residual metric below is zero. Thus $H$ need not vanish in a calibrated
law, is not convex under mixing, and has no universal ordering with binned
ECE. No population consistency or published coverage guarantee is claimed.

\paragraph{Separate residual functional.}
With $\phi_\kappa$ the Gaussian density, define
\begin{align}
 K^R_\kappa(t,c)&=\sum_{j\in\mathbb Z}
 [\phi_\kappa(t-c+2j)+\phi_\kappa(t+c+2j)],\nonumber\\
 R_\kappa(P)&=\int_0^1\left|\mathbb E_P
 K^R_\kappa(t,C)(Y-C)\right|\,dt. \label{eq:residual}
\end{align}
This is the published fixed-scale residual family, not bandwidth-free
SmoothECE, whose scale solves a fixed-point condition
\citep{blasiok2023smoothece}. We do not replace historical results by $R$.
All composition statements below use one common fixed scale.

\paragraph{Support and an independent budget.}
Let $P^\pi$ be the event/state law induced by policy $\pi$, and $U=u$
an audit cell containing state, selected branch and epoch. State is
endogenous to policy history. For eligibility $V_h$ and functional $F$,
\begin{align}
 d^F_{B:T_0,h,u}&=F(B;P^\pi\mid u,V_h)
                 -F(T_0;P^\pi\mid u,V_h),\nonumber\\
 d^F_{B:T_0,h,u}&\le b^F_{B,h,u}. \label{eq:contract}
\end{align}
$d$ is observed and signed; $b\ge0$ must be approved independently of
evaluation labels. Separate budgets are required for $W$ and $S$.
Neither a test-set difference nor a median logit perturbation is such a
budget. Empty or unsupported cells are unevaluable, not zero-error passes.
Comparing predictors on $P^{\mathrm{teacher\mbox{-}always}}$ does not
certify support selected by Schmitt. The historical Table~\ref{tab:contract}
uses precisely that common teacher trajectory, not student on-policy states.

\section{Method}
\label{sec:method}
\paragraph{LCD and replay.}
The loss combines multi-task ground-truth CE with uncertainty weighting
\citep{kendall2018multitask}, distillation \citep{hinton2015distilling}
at temperature $\tau$, load weight $\alpha(s)=0.5+1.5\hat s$, a
confidence-matching term, and an initial curriculum. The implemented
confidence term acts on gate. DAgger-style rollout/retraining
\citep{ross2011dagger} samples disagreement regions under the automaton.
This is a training recipe, not a convergence or per-stratum calibration theorem.

\paragraph{Cost-informed Schmitt design.}
The nominal width uses multiplier 1.5 times
$c_{\mathrm{sw}}/\Delta\tau_{\mathrm{ev}}$, where
$\Delta\tau_{\mathrm{ev}}>0$ is the wide--narrow service difference in
seconds/event. The ratio has units of events; normalization gives the
router's queue units. Floors and clips affect actual thresholds, and
minimum dwell is a separate parameter. The multiplier is a design choice,
not a measured optimum. Its limited scale invariance is stated below.
Historical campaign medians are 4 switches versus 420--1,541 for B4;
the reported median reduction factors are 105--385, worst paired factor
43. Counts describe transitions, not a material latency saving.
At a constant nominal 21.9\,\textmu s/switch, B4 switching costs
0.057--0.211\% of the \emph{16\,s nominal horizon}, not $\le0.2\%$ of
busy service. Wide attendance, actual batch service, staging and mapping
costs require separate accounting.

\section{Theory: Conditional Statements, Not Checkpoint Certificates}
\label{sec:theory}
The former training-improvement, transient-$H$, metric-fidelity and
DES/SLA propositions are not established and are withdrawn as guarantees.
The following replacements have full proofs in Appendix~\ref{app:theory}.

\begin{proposition}[Selected residual composition]
\label{prop:mix}
For a common nonnegative normalized kernel,
$R(\sum_u w_uP_u)\le\sum_u w_uR(P_u)\le1$.
If $P_u$ are the actual selected laws under $\pi$,
$R(T_0\mid u,V_h)\le a_u$ and
$d^R_{B(u):T_0,h,u}\le b_{B(u),h,u}$, then
\[
 R(\mathrm{served}\mid V_h)\le
 \sum_u w_u(a_u+b_{B(u),h,u}),\qquad
 w_u=\Pr(U=u\mid V_h).
\]
Conditional calibration of each selected branch is sufficient for
calibration of the served confidence. Marginal branch calibration alone
is not sufficient, even for label-blind selection.
\end{proposition}
These are law-level or empirical functional inequalities, not sampling
guarantees. Current tables establish neither the assumed budgets nor their
transfer to future selected laws. They do not imply a mixing bound for $H$.
For fixed pre/post residual laws, a window with post mass $w$ has bound
$(1-w)R_-+wR_+$. Constant-rate trailing windows lose old mass linearly as
$(1-\Delta/L)_+$, not geometrically; variable-rate streams require actual
eligible event counts. No historical transient RHS has been validated.

\begin{proposition}[Limited ratio invariance]
\label{prop:deadband}
If $c_{\mathrm{sw}}$ and $\Delta\tau_{\mathrm{ev}}>0$ are both multiplied
by the same positive factor, their ratio is unchanged. An actual
deadband proportional to this ratio is unchanged only where dimensional
floors are inactive and the normalization/clipping rule is unchanged.
This does not transport dwell, arrivals, the SLA or a state trajectory.
\end{proposition}

\begin{proposition}[Pure Schmitt switch count]
\label{prop:switch}
For fixed thresholds separated by $d_q>0$, signal total variation $V_q$,
horizon $H$ and minimum dwell $\tau_d>0$, with no resets or overrides
bypassing crossings,
$N_{\mathrm{sw}}\le1+V_q/d_q$ and
$N_{\mathrm{sw}}\le1+H/\tau_d$.
\end{proposition}
The DES's guard behavior must satisfy these hypotheses before applying
the count bound. It is not a guarantee of measured p99 latency.

\begin{proposition}[Periodic fluid sufficiency]
\label{prop:fluid}
For a work-conserving, lossless reflected fluid queue with integrable,
deterministic $T$-periodic arrival/capacity rates $\lambda,\mu\ge0$, include
switch pauses as $\mu=0$. If
$\int_0^T(\mu-\lambda)dt>0$, every finite initial backlog eventually
coincides with a bounded periodic trajectory. A negative integral gives
unbounded backlog; equality need not do so. This is not a theorem for the
random-batch DES or a percentile SLA.
\end{proposition}

\section{Experiments and Provenance}
\label{sec:exp}
The extended campaign has $\gamma\in\{0,0.25,0.5,1\}$ and seeds 0--9
(40 cells, 16 rows/cell). The original campaign, ablations and router
equalization used seeds 0--2; later extensions and sensitivities are not
retroactively pre-registered. Service RNG seed 7420 resets per deployment.
Historical transient redeployment reproduced all 40 Schmitt serving rows
to $10^{-6}$; this checks reproduction, not statistical validity or
equality of per-event shocks across policies with different batches.

Teacher violation is 70.6--70.8\%; student 1.5--1.8\% (p99 1.6--1.7\,ms,
p50 0.7\,ms); Schmitt 1.4--4.3\% (p99 1.6--13\,ms), coverage 0.67--0.70,
typically 4 switches. These summaries do \emph{not} meet the literal
p99 1.5\,ms SLA. LCD and LCD+DAgger Schmitt serving rows coincide at this
operating point; no additional serving margin is established by that equality.

\begin{table*}[t]
\centering\small\begin{tabular}{llrrrr}
\toprule
$\gamma$ & Method & viol\% & p99\,ms & switches & cov \\
\midrule
$\gamma0$ & B1: teacher (saturated) & 70.6$\pm$(0.2) & 168.3 & 0 & 0.00 \\
 & B1': student-only (LCD) & 1.8$\pm$(0.6) & 1.6 & 0 & 1.00 \\
 & Schmitt·LCD+DAgger (ours) & 4.3$\pm$(5.2) & 13.0 & 4 & 0.67 \\
 & Schmitt·LCD (ours) & 4.3$\pm$(5.2) & 13.0 & 4 & 0.67 \\
 & B4: fixed-threshold (best-q) & 1.9$\pm$(0.6) & 1.7 & 1,279 & 0.69 \\
 & B5: random matched & 64.4$\pm$(4.0) & 24.3 & 16,676 & 0.68 \\
\addlinespace
$\gamma0.25$ & B1: teacher (saturated) & 70.6$\pm$(0.1) & 165.0 & 0 & 0.00 \\
 & B1': student-only (LCD) & 1.6$\pm$(0.9) & 1.7 & 0 & 1.00 \\
 & Schmitt·LCD+DAgger (ours) & 3.1$\pm$(4.8) & 8.8 & 4 & 0.69 \\
 & Schmitt·LCD (ours) & 3.1$\pm$(4.8) & 8.8 & 4 & 0.69 \\
 & B4: fixed-threshold (best-q) & 1.6$\pm$(0.7) & 1.6 & 1,335 & 0.70 \\
 & B5: random matched & 64.4$\pm$(4.6) & 17.3 & 16,324 & 0.69 \\
\addlinespace
$\gamma0.5$ & B1: teacher (saturated) & 70.7$\pm$(0.2) & 166.8 & 0 & 0.00 \\
 & B1': student-only (LCD) & 1.7$\pm$(0.5) & 1.6 & 0 & 1.00 \\
 & Schmitt·LCD+DAgger (ours) & 2.1$\pm$(1.1) & 2.7 & 4 & 0.70 \\
 & Schmitt·LCD (ours) & 2.1$\pm$(1.1) & 2.7 & 4 & 0.70 \\
 & B4: fixed-threshold (best-q) & 1.8$\pm$(0.6) & 1.7 & 856 & 0.69 \\
 & B5: random matched & 65.1$\pm$(3.6) & 15.5 & 16,280 & 0.70 \\
\addlinespace
$\gamma1$ & B1: teacher (saturated) & 70.6$\pm$(0.2) & 168.9 & 0 & 0.00 \\
 & B1': student-only (LCD) & 1.5$\pm$(0.6) & 1.6 & 0 & 1.00 \\
 & Schmitt·LCD+DAgger (ours) & 1.4$\pm$(0.6) & 1.6 & 4 & 0.70 \\
 & Schmitt·LCD (ours) & 1.4$\pm$(0.6) & 1.6 & 4 & 0.70 \\
 & B4: fixed-threshold (best-q) & 1.5$\pm$(0.7) & 1.6 & 796 & 0.70 \\
 & B5: random matched & 62.2$\pm$(5.7) & 16.8 & 16,173 & 0.70 \\
\bottomrule
\end{tabular}

\caption{Extended campaign serving summary, 10 seeds per $\gamma$.
Violation is the fraction above 1.5\,ms; displayed violation is mean
$\pm$ seed sd, p99/switches/coverage are seed means. B4 thresholds were
selected on validation per cell, then frozen for test; no test reselection
is implied. Separately quoted 420--1,541 counts are campaign medians.
The historical equalization sweep has only 12 cells.}
\label{tab:serving}
\end{table*}

\begin{figure*}[t]
\centering\includegraphics[width=0.92\textwidth]{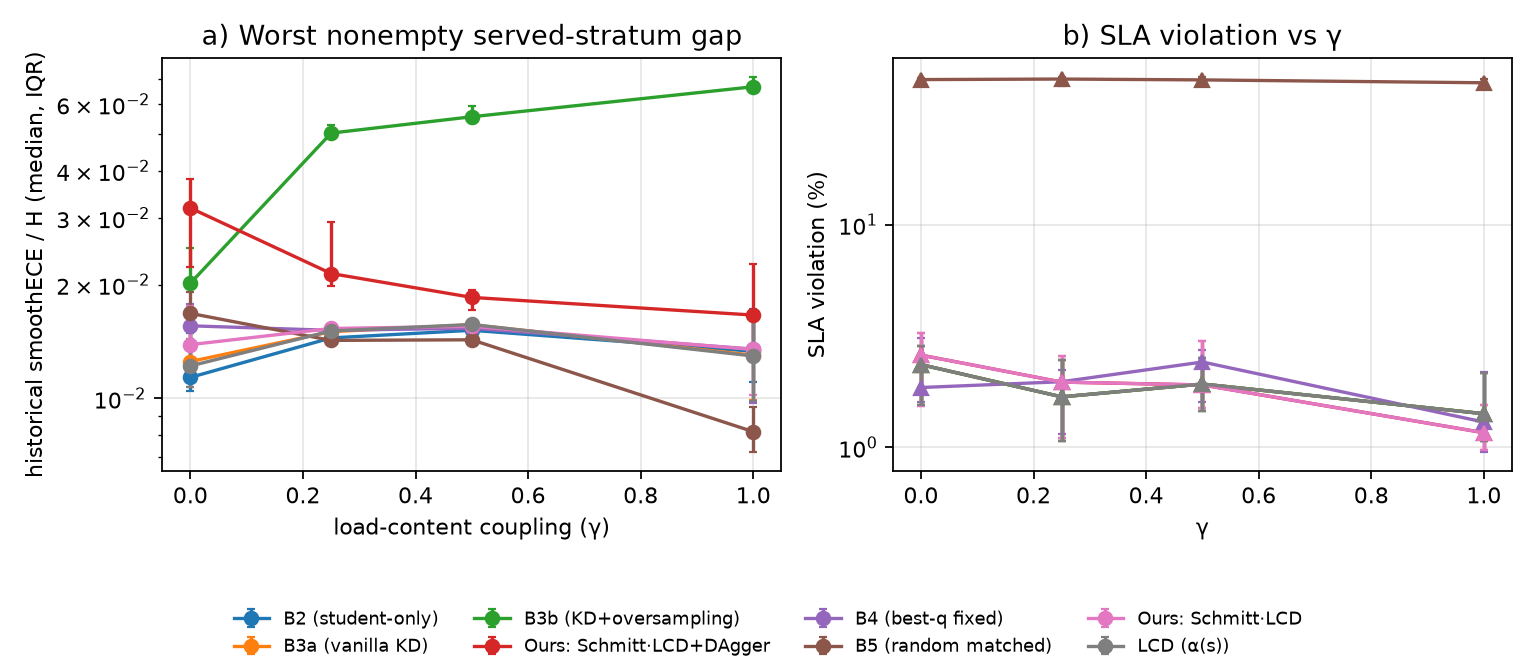}
\Description{Two panels show median historical kernel correctness gap and
SLA violation by coupling, with descriptive seed interquartile ranges.}
\caption{Campaign summaries versus $\gamma$: medians and descriptive
seed IQRs, not standard errors or confidence intervals. Left: worst
nonempty served-stratum $H_\kappa$ (log scale); right: SLA violation
(symlog). Plot label ``smoothECE'' is a legacy identifier for $H_\kappa$.}
\label{fig:gamma}
\end{figure*}

\paragraph{Training-protocol calibration.}
Table~\ref{tab:g2} reports hot-stratum $H$ on glass-box states with fitted
temperatures. LCD/B3a corn reductions are +6.5/+2.0/$-$0.9/$-$5.8\%
by $\gamma$; seed-level favorables 18/40. The $\ge30\%$ criterion against
vanilla KD fails. LCD/B3b corn reductions are 40--77\%; this is a
comparison with a damaged blind-oversampling baseline, not general
superiority of load coupling. At $\gamma=0$, reported gate/choice mean
differences are within 0.01\,pts and corn mean difference is +0.02\,pts
(4/10 favorable, maximum absolute seed difference 0.20\,pts).
These descriptive differences lie within the proposed $\pm0.3$ margin;
without the corresponding two-sided equivalence test/intervals for the
declared family, we do not label every head or stratum TOST-equivalent.

\begin{figure*}[t]
\centering\includegraphics[width=0.85\textwidth]{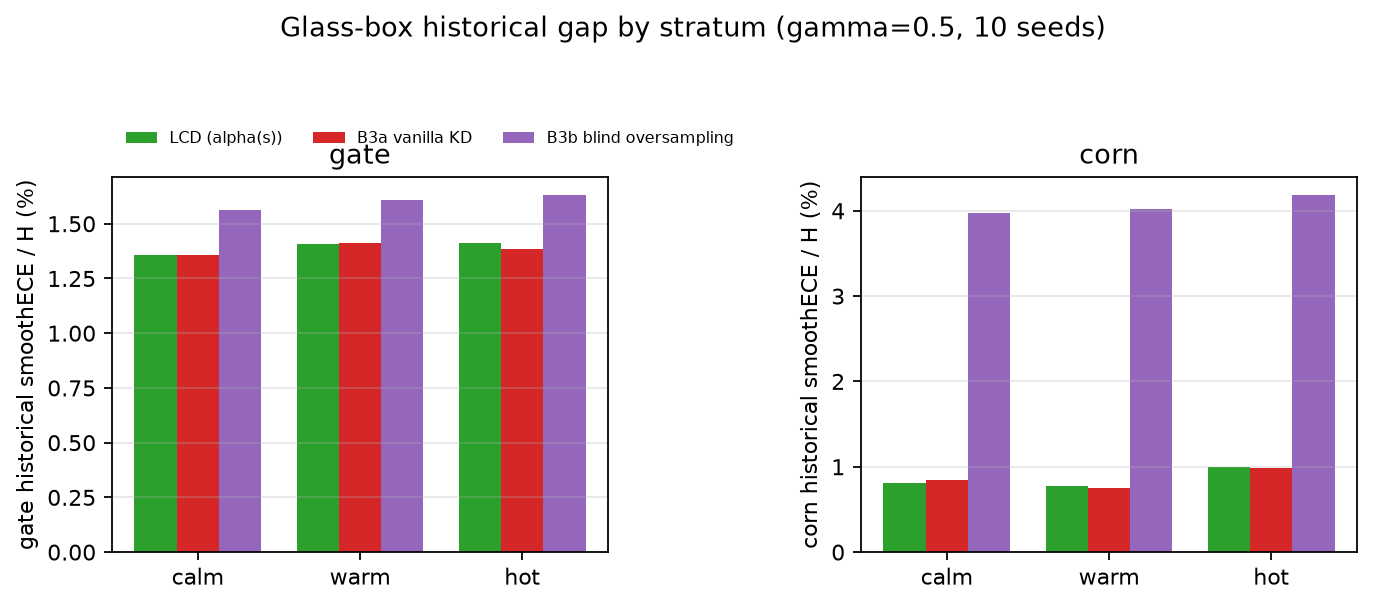}
\Description{Gate and ordinal historical gaps in calm, warm and hot
teacher-trajectory strata, comparing LCD, vanilla KD and blind oversampling.}
\caption{Glass-box training-protocol gap by stratum at $\gamma=0.5$
(10-seed medians). ``smoothECE'' denotes historical $H_\kappa$.
Gate differences are descriptively small; corn separates KD/LCD from
B3b. No every-stratum equivalence test is asserted. Student-only deployed
warm/hot strata are empty and cannot support this comparison.}
\label{fig:strata}
\end{figure*}
\begin{table*}[t]
\centering\small\begin{tabular}{llrrrrr}
\toprule
$\gamma$ & head & LCD & B3a & B3b & LCD/B3a & LCD/B3b \\
\midrule
$\gamma0$ & gate & 1.09 & 1.08 & 1.21 & -0.8\% & +10.3\% \\
 & choice & 1.13 & 1.13 & 1.31 & -0.2\% & +13.8\% \\
 & corn & 2.12 & 2.27 & 3.56 & +6.5\% & +40.4\% \\
\addlinespace
$\gamma0.25$ & gate & 1.54 & 1.52 & 1.72 & -1.3\% & +10.6\% \\
 & choice & 1.66 & 1.66 & 1.97 & +0.1\% & +15.6\% \\
 & corn & 1.22 & 1.24 & 4.73 & +2.0\% & +74.2\% \\
\addlinespace
$\gamma0.5$ & gate & 1.41 & 1.39 & 1.63 & -1.9\% & +13.5\% \\
 & choice & 1.58 & 1.60 & 1.91 & +1.0\% & +17.2\% \\
 & corn & 0.99 & 0.98 & 4.19 & -0.9\% & +76.3\% \\
\addlinespace
$\gamma1$ & gate & 0.00 & 0.00 & 0.00 & -- & -- \\
 & choice & 0.43 & 0.43 & 0.46 & -2.0\% & +4.6\% \\
 & corn & 0.97 & 0.92 & 4.27 & -5.8\% & +77.2\% \\
\bottomrule
\end{tabular}

\caption{Hot-stratum training-protocol $H_\kappa$, $\times100$, medians
over 10 seeds; historical ``smoothECE'' identifier. Relative reductions
are ratios of medians, not medians of paired reductions. Gate at
$\gamma=1$ is degenerate (all attacks).}
\label{tab:g2}
\end{table*}

\paragraph{Exploratory ablations.}
The three-seed ablation grid, not rerun at ten seeds, gives corn hot
2.83/2.82 versus LCD 2.83\,pts at $\gamma=0$ when confidence/curriculum
terms are removed, and 0.89/0.85 versus 0.90 at $\gamma=0.5$.
DAgger oversampling=1 gives 0.90 versus 0.89 at $\gamma=0.5$.
These ablations did not detect an additional effect in these pools;
they do not identify $\alpha$ as the sole mechanism or exonerate all
oversampling. Masks, effective sample counts and optimization steps are
not factorially matched; the B3b implementation adds one copy of error
indices, not a universal threefold weight. A no-state-input ablation is
not part of this implemented set.

\begin{figure}[t]
\centering\includegraphics[width=0.72\linewidth]{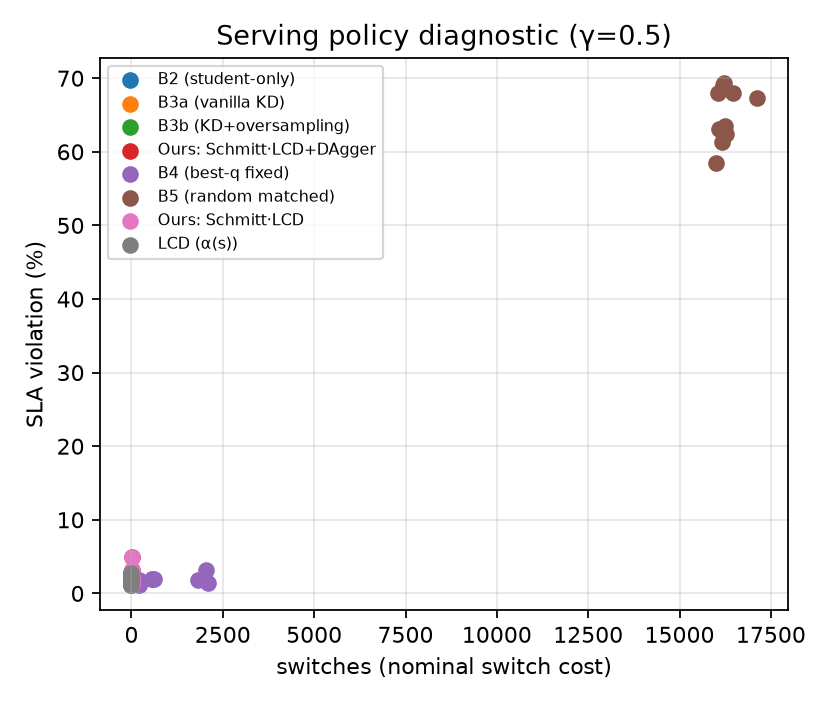}
\Description{Campaign scatter of switches against latency violations,
showing policy-specific tails rather than certified Pareto superiority.}
\caption{Campaign transitions versus violation at $\gamma=0.5$.
Many fewer transitions do not establish a latency Pareto gain: nominal
B4 switch time is 0.057--0.211\% of the horizon across $\gamma$, and
the switched violation tail remains important.}
\label{fig:pareto}
\end{figure}

\paragraph{Historical router equalization.}
The 12-cell sweep (seeds 0--2) reports sealed-point violation differences
versus validation-selected B4 of +0.30/+0.09/+0.05\,pts at
$\gamma=0,0.25,1$. At $\gamma=0.5$, the sealed point is 1.2\,pts outside
the $\pm1$\,pt band; inspecting the test sweep gives $-$0.15\,pts at
\emph{multiplier} 0.85, meaning $q_{\mathrm{on}}=0.85\times0.675
\approx0.574$, not threshold 0.85. This is a retrospective sweep
descriptor, not an independently validated replacement policy.
Reported switches are 4--6 versus 200--2,044 (historical selector output);
zero-off-threshold candidates are excluded as effectively always-student.
B4 is not reselected on test by the equalization script. The 40-cell
student-only comparison below is a different comparator and cannot
approve the sweep's B4 criterion.

\paragraph{Transient descriptors.}
Historical F5 covers 40 cells and 163 windows $\pm2$\,s around an
\emph{arrival-time proxy} for the first switch-affected event, not an
observed service-start timestamp. Bootstrap bands use 200 deterministic
draws of event-count blocks, not 1,000 hierarchical time-block draws.
The 30\% non-overlap of narrow-head pre/post 95\% bands is descriptive,
not a paired or multiplicity-adjusted jump test. Absolute differences
are at most 0.6\,pts at $\gamma=0$ and 1.0 at $\gamma=0.5$; $\gamma=1$
has degenerate gate behavior. Corn posts span 0.9--17.8\,pts, drain
medians 3.6--5.1, delta medians +0.8 to +3.6 versus glass-box hot levels
0.8--2.2. These regimes differ in selected population and state; no
transient bound for $H$ has been validated by their bands.

\begin{figure*}[t]
\centering\includegraphics[width=0.92\textwidth]{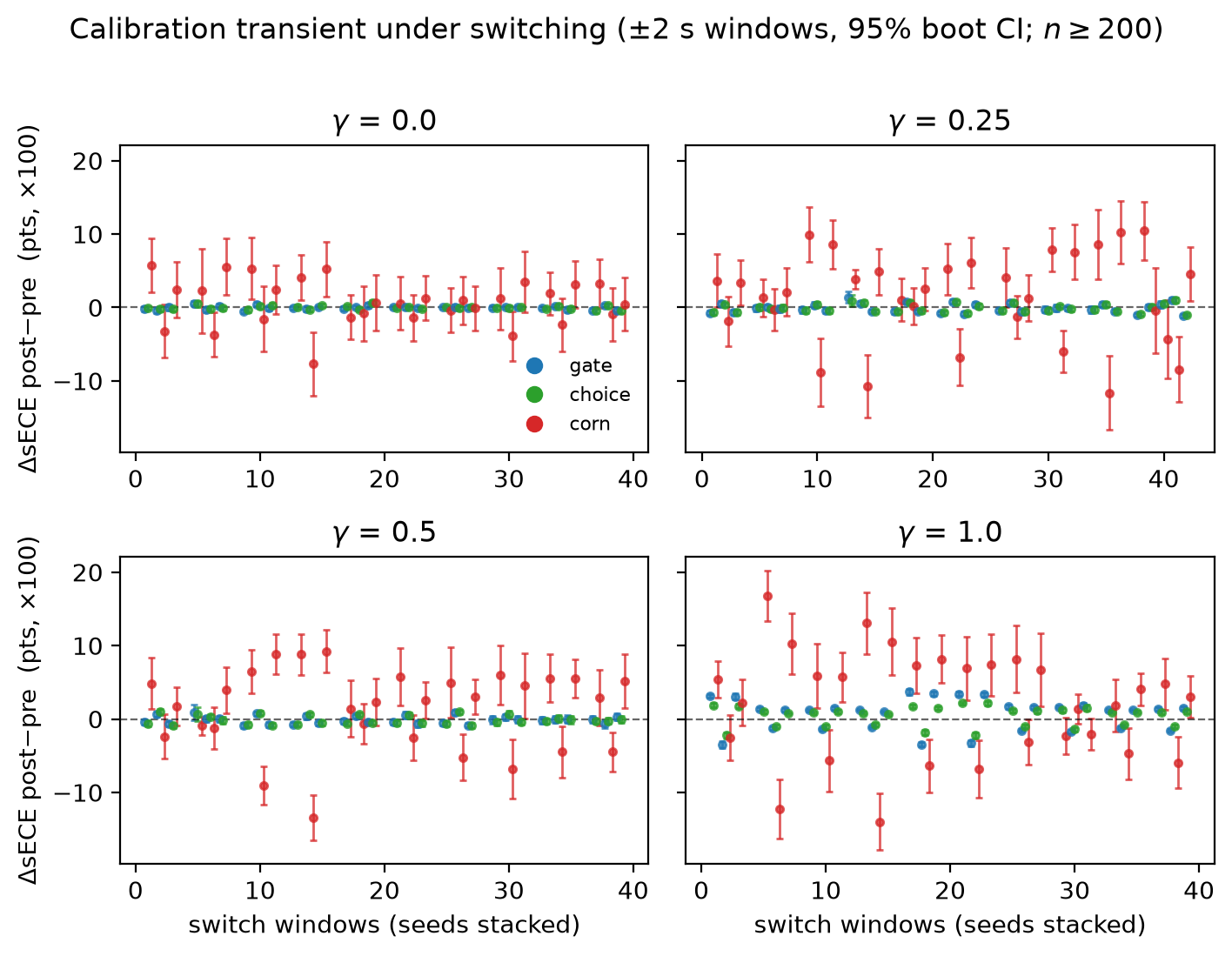}
\Description{Four coupling panels of historical post-minus-pre kernel
correctness gaps with descriptive event-block bootstrap bands.}
\caption{Historical post-minus-pre $H_\kappa$ (legacy ``smoothECE''),
40 cells/163 proxy-centered windows. Displayed 95\% bootstrap bands
are descriptive, based on 200 event-block draws; their coverage under
dependence/selection is unvalidated. They do not test a predictive RHS.}
\label{fig:transient}
\end{figure*}

\subsection{Common-Trajectory and Served-Stream Diagnostics}
\label{sec:f2}
\paragraph{Common teacher-trajectory audit, not student on-policy.}
Table~\ref{tab:contract} recomputes $T_0$ and counterfactual LCD+DAgger
$S$ on the same teacher-always replay states, using serving transformations
without temperatures. The serving-protocol teacher recomputation matches
its recorded values to $5\times10^{-5}$; the separate tempered audit
matches the campaign warm/hot values to $3\times10^{-5}$.
Over 40 cells, median absolute observed deficit is 0.06\,pts:
334/360 entries are within 0.30, 354/360 within 0.67, maximum 1.9
at $\gamma=0$ hot corn (about 0.9k events, seed 8).
Accuracy differences are within 0.3\,pts in 338/360 entries, maximum
1.8. These are original cell-level counts, not deficits on Schmitt's
student-selected support; small accuracy differences do not establish
quality or calibration equivalence. No independent $b$ is tested here.
The displayed table difference is a difference of seed medians, whereas
the 334/360 count uses paired deficits per seed.

\begin{table*}[t]
\centering\footnotesize\begin{tabular}{llrrrrrr}
\toprule
$\gamma$ & state.head & $T_0$-acc & S-acc & $T_0$-H & S-H & $d$ (pts) & $n$ \\
\midrule
$\gamma0$ & calm/gate & 97.1 & 97.1 & 1.02 & 1.20 & 0.18 & 100758 \\
 & calm/choice & 97.1 & 97.1 & 1.07 & 1.23 & 0.16 & 100758 \\
 & calm/corn & 67.0 & 66.4 & 1.73 & 1.58 & -0.15 & 100758 \\
\addlinespace
 & warm/gate & 97.1 & 97.1 & 1.04 & 1.04 & -0.00 & 77072 \\
 & warm/choice & 97.1 & 97.1 & 1.07 & 1.07 & -0.00 & 77072 \\
 & warm/corn & 67.0 & 66.6 & 1.52 & 1.67 & 0.15 & 77072 \\
\addlinespace
 & hot/gate & 97.1 & 97.1 & 1.08 & 1.07 & -0.01 & 19392 \\
 & hot/choice & 97.1 & 97.1 & 1.11 & 1.10 & -0.01 & 19392 \\
 & hot/corn & 67.8 & 68.0 & 2.40 & 2.41 & 0.01 & 19392 \\
$\gamma0.25$ & calm/gate & 94.7 & 94.7 & 1.30 & 1.46 & 0.16 & 98520 \\
 & calm/choice & 94.7 & 94.7 & 1.34 & 1.44 & 0.10 & 98520 \\
 & calm/corn & 67.2 & 67.1 & 1.03 & 1.10 & 0.07 & 98520 \\
\addlinespace
 & warm/gate & 90.9 & 90.9 & 1.34 & 1.38 & 0.04 & 80296 \\
 & warm/choice & 90.9 & 90.9 & 1.42 & 1.45 & 0.03 & 80296 \\
 & warm/corn & 67.2 & 67.2 & 0.75 & 0.78 & 0.03 & 80296 \\
\addlinespace
 & hot/gate & 90.9 & 90.9 & 1.48 & 1.46 & -0.02 & 19016 \\
 & hot/choice & 90.9 & 90.9 & 1.60 & 1.65 & 0.05 & 19016 \\
 & hot/corn & 67.6 & 67.4 & 1.13 & 1.05 & -0.08 & 19016 \\
$\gamma0.5$ & calm/gate & 93.9 & 93.9 & 1.34 & 1.44 & 0.10 & 97709 \\
 & calm/choice & 93.9 & 93.9 & 1.36 & 1.44 & 0.08 & 97709 \\
 & calm/corn & 67.2 & 67.2 & 0.87 & 0.83 & -0.04 & 97709 \\
\addlinespace
 & warm/gate & 88.9 & 88.9 & 1.34 & 1.47 & 0.13 & 79480 \\
 & warm/choice & 88.9 & 88.9 & 1.43 & 1.53 & 0.09 & 79480 \\
 & warm/corn & 67.2 & 67.2 & 0.86 & 0.83 & -0.03 & 79480 \\
\addlinespace
 & hot/gate & 88.9 & 88.8 & 1.40 & 1.42 & 0.01 & 23104 \\
 & hot/choice & 88.8 & 88.8 & 1.53 & 1.49 & -0.04 & 23104 \\
 & hot/corn & 67.2 & 67.2 & 0.99 & 1.02 & 0.03 & 23104 \\
$\gamma1$ & calm/gate & 98.2 & 98.2 & 0.77 & 0.78 & 0.02 & 100475 \\
 & calm/choice & 97.8 & 97.8 & 0.84 & 0.87 & 0.03 & 100475 \\
 & calm/corn & 67.0 & 67.1 & 0.81 & 0.83 & 0.02 & 100475 \\
\addlinespace
 & warm/gate & 100.0 & 100.0 & 0.00 & 0.00 & 0.00 & 80304 \\
 & warm/choice & 99.1 & 99.1 & 0.39 & 0.40 & 0.01 & 80304 \\
 & warm/corn & 67.2 & 67.2 & 0.71 & 0.68 & -0.03 & 80304 \\
\addlinespace
 & hot/gate & 100.0 & 100.0 & 0.00 & 0.00 & 0.00 & 20208 \\
 & hot/choice & 99.1 & 99.1 & 0.43 & 0.43 & 0.00 & 20208 \\
 & hot/corn & 67.3 & 67.3 & 0.93 & 0.89 & -0.03 & 20208 \\
\bottomrule
\end{tabular}

\caption{Common teacher-trajectory audit ($P^{\mathrm{teacher\mbox{-}always}}$),
medians over 10 seeds, $\times100$, no temperatures. $T_0$ versus
counterfactual LCD+DAgger $S$; legacy filename/``sECE'' denote $H_\kappa$.
Displayed $d=$ difference of medians, not an independent budget.
$n$ is stream events per stratum, not the corn-eligible count. This is
not an on-policy contract for the deployed student.}
\label{tab:contract}
\end{table*}

\paragraph{Fixed-state predictor counterfactual.}
Wide-mode attendance covers about 30\% of events, chiefly idle/calm
phases, and 30.3/2.1\% of corn at $\gamma=0/1$.
Both student versions are replayed on those actual Schmitt events and
states. Student-served DES/batch confidence differences are at most
$8\times10^{-7}$ with 100\% argmax agreement.
The \emph{served wide branch} has median accuracy differences versus
counterfactual students of $-$1.3 to $-$2.6\,pts for narrow heads and
$-$1.1 to $-$12.7 for corn (extremes $-$23.5); one favorable entry
occurs among $40\times3$ cell/head pairs (+0.9\,pts, $\gamma=0$ corn).
The two student versions differ by at most 0.6\,pts at medians and
0.9 per seed. This identifies an accuracy comparison on a fixed slice,
not a general no-quality-gap result or a causal calibration decomposition.

\paragraph{Original teacher versus the deployed wide branch.}
Wide-branch corn accuracy medians are 64.8/59.8/57.0/53.4\%, versus
$T_0$ 66.9/66.8/67.0/66.9 on the same attended events and student
65.9--67.0. $T_0$ beats $W$ in 40/40 cells by median 1.7--12.4\,pts.
The historical branch probe verifies zero \emph{argmax disagreement}
between serving and checkpoint-wide decisions in 40/40 cells; it does
not establish identical probabilities. Architecture establishes $W$'s
identity, but this probe does not measure $T_0$ probability calibration.
Paired $T_0-S$ slice-accuracy differences have medians
+0.700/$-$0.033/$-$0.099/$-$0.229\,pts by $\gamma$; no general utility
or calibration equivalence is inferred from these values or from
Table~\ref{tab:contract}, which uses a different trajectory.
Serving $T_0$ itself and measuring its real costs remains future work.

\begin{table*}[t]
\centering\footnotesize\begin{tabular}{lrrrrrrrrr}
\toprule
& \multicolumn{3}{c}{gate} & \multicolumn{3}{c}{choice} & \multicolumn{3}{c}{corn (attacks)} \\
\cmidrule(lr){2-4}\cmidrule(lr){5-7}\cmidrule(lr){8-10}
$\gamma$ & W & cf-S & cf-L & W & cf-S & cf-L & W & cf-S & cf-L \\
\midrule
$\gamma0$ & 94.5 & 97.1 & 97.1 & 94.7 & 97.1 & 97.1 & 64.8 & 65.9 & 66.9 \\
$\gamma0.25$ & 95.4 & 97.1 & 97.1 & 95.6 & 97.1 & 97.1 & 59.8 & 66.8 & 67.1 \\
$\gamma0.5$ & 95.2 & 97.1 & 97.1 & 95.8 & 97.1 & 97.1 & 57.0 & 67.0 & 67.3 \\
$\gamma1$ & 95.2 & 97.1 & 97.1 & 95.6 & 97.1 & 97.1 & 53.4 & 66.7 & 66.8 \\
\bottomrule
\end{tabular}

\caption{Accuracy on actual wide-served events/states, 10-seed medians,
$\times100$. $W$ is the frozen wide head on adapted trunk, not $T_0$;
cf-S is LCD+DAgger, cf-L is LCD on the same slice. Accuracy does not
certify probability/calibration equivalence.}
\label{tab:quality}
\end{table*}

\paragraph{Two student-only comparators, two estimands.}
Campaign B1$'$ uses \emph{LCD}; F2 Schmitt uses \emph{LCD+DAgger}.
LCD student-only has lower full-stream corn $H$ in 38/40 cells:
10/10,10/10,10/10,8/10 by $\gamma$, with reversals at $\gamma=1$ seeds
1 and 4. Its medians are 1.22/0.81/0.82/1.29\,pts versus switched
\emph{levels} 3.19/2.14/1.71/1.66, not gaps.
H5 instead uses \emph{the same LCD+DAgger checkpoint} for student-only
and Schmitt: student-only has lower corn $H$ in 27/40 cells
(8/10,8/10,6/10,5/10). Neither comparison is universal success,
and the first also changes checkpoints.
F2 paired median violation differences versus campaign student-only are
within 0.1\,pts, but Schmitt has three tails above 10\%:
$\gamma=0$ seeds 5/8 at 12.228/16.667\%, and $\gamma=0.25$ seed 3
at 17.460\%; queue maxima are 1.6--2.5k, while student-only is
$\le3.5\%$ on these traces. Median parity hides this risk.

\paragraph{Weighted slice diagnostic, not additive decomposition.}
The historical product of eligible wide-slice share and
$(H_W-H_{S,\mathrm{cf}})$ has medians 1.12/0.39/0.14/0.10\,pts
(range $-$0.7 to +1.7). Wide-slice corn $H$ medians are 5.3--6.6
(range 0.7--13.6), versus counterfactual students 1.6--2.0 (1.0--3.1).
These are useful local descriptors. Because the kernel fit and absolute
value depend on the full population, the product is \emph{not} the global
effect of replacing $W$ by $S$. The recorded 3.19/2.14/1.71/1.66 are
full-stream levels; subtracting the product cannot recover a causal
``remainder.'' A correct fixed-trajectory global counterfactual requires
recomputing the metric after replacing all wide predictions, and remains
pending. Appendix~\ref{app:theory} gives a valid telescoping identity,
not an identified unique mediation.

\paragraph{Small-pool sensitivity.}
Historical controls use 200 matched $n=500$ draws per cell.
Corn drain windows exceed their own stream's 97.5th percentile in
group counts 4/20,11/20,12/20,14/20, totaling 41/80, versus 1/42
student-burst comparisons. An independent recheck of the recorded groups
corrects the earlier textual denominator 70 to 80; no exclusions or window
values were changed. These exceedance counts are descriptive, not tests.
Wide-served drain medians are 5.5--7.8\,pts. Corn farther than
$\pm2$\,s from switches comprises 937--1741 of about 76k events
(1.2--2.3\%), entirely wide-served, with $n=500$ bands 1.6--17.1\,pts.
These are descriptive regime/sample-size sensitivities, not multiple
hypothesis tests or evidence of a clean served steady-state corn pool.

\paragraph{Retrospective temporal recalibration.}
H4 fits correctness maps on arrivals in the first half ($t_{\mathrm{arr}}
\le8$\,s), excluding 1\,s warmup, and evaluates the second half.
The split aligns nominal burst duty, not necessarily mode masses or
label availability. It does \emph{not} filter
$t_{\mathrm{dep}}+\tau_L\le8$\,s. Pending final/fit labels cannot be
counted from the saved aggregate artifacts; causal delayed-label
recalibration has not been implemented/validated by these numbers.
Platt maps $\sigma(a\operatorname{logit}c+b)$ output correctness
probabilities for fixed chosen labels, not a coherent multiclass PMF.
They are outside the DES clock, not measured zero-cost operations.

Corn evaluation-half $H$ medians fall $2.95\to1.64$,
$2.03\to0.91$, $1.65\to0.96$ and $1.68\to0.78$\,pts; 39/40 improve,
one $\gamma=0$ seed loses 0.9\,pts from 2.1. Mixture maps are within
0.3\,pts of mode maps at per-$\gamma$ medians.
Slopes 0.94--1.43 in student mode and up to 2.5 in wide mode do not,
without intercept/domain analysis, prove identity or a causal need for
free slopes. The historical intercept-only baseline is not a properly
optimized MLE; a correct slope-one fit remains pending.
Student-only LCD+DAgger with its own map reaches 1.88/0.97/0.91/0.78.
Retrospectively declared paired four-$\gamma$ families (ten seeds/group)
give marginal 90\% t intervals for mode-repaired switched minus own-map
student-only of [$-$0.263,+0.146], [$-$0.085,+0.053],
[$-$0.008,+0.092], [$-$0.014,+0.037]\,pts, all Holm-adjusted
two-sided $p\ge0.6348$. Mode-versus-mixture shows no incremental benefit
detected after Holm (smallest adjusted $p=0.176$). These are neither
simultaneous intervals nor equivalence tests; they do not remove the
availability caveat or show that the entire quality gap is repaired.

\begin{table*}[t]
\centering\footnotesize\begin{tabular}{lrrrrr}
\toprule
$\gamma$ & served & mixt.-Platt & mode-Platt & mode-iso & same-S+recal \\
\midrule
$\gamma0$ & 2.95 {\scriptsize(0.71--4.45)} & 1.92 {\scriptsize(0.95--2.79)} & 1.64 {\scriptsize(0.70--2.94)} & 2.28 {\scriptsize(1.12--3.45)} & 1.88 {\scriptsize(1.17--2.42)} \\
$\gamma0.25$ & 2.03 {\scriptsize(1.35--3.16)} & 1.04 {\scriptsize(0.77--3.99)} & 0.91 {\scriptsize(0.73--1.14)} & 1.18 {\scriptsize(0.78--1.79)} & 0.97 {\scriptsize(0.65--1.13)} \\
$\gamma0.5$ & 1.65 {\scriptsize(0.94--2.98)} & 1.00 {\scriptsize(0.79--1.28)} & 0.96 {\scriptsize(0.73--1.11)} & 1.25 {\scriptsize(0.72--1.47)} & 0.91 {\scriptsize(0.68--1.09)} \\
$\gamma1$ & 1.68 {\scriptsize(0.86--3.91)} & 0.79 {\scriptsize(0.68--1.21)} & 0.78 {\scriptsize(0.63--1.01)} & 0.86 {\scriptsize(0.75--1.14)} & 0.78 {\scriptsize(0.66--1.01)} \\
\bottomrule
\end{tabular}

\caption{Retrospective temporal correctness recalibration, evaluation-half
$H_\kappa$ (legacy ``sECE\_eval''), 10-seed medians with ranges.
Arrival-filtered fits do not establish causal delayed-label availability.
Last column is same LCD+DAgger student-only with its own map, not campaign
LCD B1$'$. No universal equivalence or zero serving cost is asserted.}
\label{tab:recal}
\end{table*}

\paragraph{Policy interventions: total DES effects, not full mediation.}
H5 retains arrivals, checkpoint and service-clock configuration, but
policy changes generate different states and realized batches; the
batches are not held fixed. Deadband-off-zero produces one switch and
then student-only, reproducing same-net student-only corn $H$ within
0.018\,pts in 40/40 cells. This is a harness-consistency check.
Same-checkpoint student-only medians are 2.01/1.83/1.58/1.58 versus
Schmitt 3.19/2.14/1.71/1.66; violation medians differ within 0.2\,pts.
The 27/40 per-cell direction above limits any universal interpretation.
Observed-load jitter held for 100\,ms, at $\pm5\%$ of $q_{\mathrm{on}}$,
reassigns individual cells to/from the queue-explosion basin (violation
changes up to 15\,pts, corn changes within 0.3\,pts); $\pm15\%$ gives
corn changes within 1.1\,pts. Delayed observed load by 0.5--2\,s gives
40--70\% violation and much larger queues. These interventions establish
sensitivity to load feedback within this DES. They do not show complete
causal attribution of the ordinal deficit to crossings, a certified
fallback outside student support or guaranteed utility of a true teacher.

\begin{table*}[t]
\centering\footnotesize\begin{tabular}{lrrrrrrrrr}
\toprule
 & sw & \multicolumn{2}{c}{$\gamma0$} & \multicolumn{2}{c}{$\gamma0.25$} & \multicolumn{2}{c}{$\gamma0.5$} & \multicolumn{2}{c}{$\gamma1$} \\
 &  & viol\% & corn  & viol\% & corn  & viol\% & corn  & viol\% & corn  \\
\midrule
b1\_samenet & 0 & 1.84 & 2.01 & 1.51 & 1.83 & 1.64 & 1.58 & 1.34 & 1.58 \\
asis & 4 & 1.94 & 3.19 & 1.66 & 2.14 & 1.63 & 1.71 & 1.15 & 1.66 \\
jit5 & 4 & 1.94 & 3.19 & 1.81 & 2.44 & 1.63 & 1.79 & 1.15 & 1.66 \\
jit15 & 4 & 1.95 & 3.14 & 2.10 & 2.56 & 1.87 & 1.79 & 1.42 & 1.86 \\
delay0.5 & 4 & 38.17 & 2.94 & 39.70 & 2.61 & 39.65 & 2.21 & 39.94 & 2.55 \\
delay2.0 & 4 & 70.51 & 2.69 & 70.45 & 4.76 & 70.52 & 4.06 & 70.49 & 5.14 \\
db\_max & 1 & 1.84 & 2.01 & 1.55 & 1.83 & 1.63 & 1.58 & 1.32 & 1.58 \\
\bottomrule
\end{tabular}

\caption{H5 policy interventions, 10-seed medians per $\gamma$;
full-stream historical $H_\kappa$. Same LCD+DAgger checkpoint and arrival
stream; policies induce different states/batches. ``b1\_samenet'' is
same-checkpoint student-only; jitter shares a per-seed noise trace;
``db\_max'' has one initial crossing then student-only.}
\label{tab:interv}
\end{table*}

\section{Scorecard, Monitoring and Limitations}
\label{sec:limits}
\paragraph{Retrospective CUSUM curve.}
The monitor compares windowed corn $H$ with a slot-specific clean median,
not an independently approved branch budget. $W=4$\,s, label delay
1\,s, step 0.25\,s and slack $k=0.01$ in fractional-gap units.
The discrete CUSUM sum is not multiplied by step; thresholds $h$ below
are in cumulative fractional-gap update units, not seconds or gap points.
Baseline includes the evaluated seed, and $h$ is scanned on the same
clean traces. Overlapping windows are strongly dependent; slot-specific
reference construction uses the full clean horizon. All saved slots here
have at least 500 events; forward-hold of invalid windows remains a
code-path risk, not an observed invalid-slot loss in these results.

D1 at $H/3$ swaps the \emph{entire B3b checkpoint}, including $W/S$,
not a smaller architecture or a student-only intervention.
At $h=0.3$, five historical false alarms accompany 30/40 detections,
median delay 6.2\,s \emph{among detected cells}; by $\gamma$ counts are
0/10,10/10,10/10,10/10. At $h=0.8$, zero historical false alarms
accompany 13/40 detections, detected median 8.4\,s. Misses and final-label
censoring prevent these medians being an unconditional reaction time.
Minimum detected delays are 3.42\,s at $h=0.3$ and 1.92 at $h=0.1$:
$W+\tau_L=5$\,s is time to full postchange exposure, not a detection floor.
D2 (duty 0.25$\to$0.60, new arrival seed and reset burst phase) has
3/40 detections at $h=0.1$, 1/40 at 0.2 and zero at 0.3/0.8;
alarm counts do not establish actual budget breaches.
Nominal horizons total 640\,s, monitored first-to-last spans 390\,s,
and 1,600 updates total 400 slot-seconds (5--14.75\,s per cell).
These are distinct exposure conventions, not independent monitoring time.
Leave-one-seed-out baseline sensitivity at fixed $h=0.3$ raises false
alarms from 5 to 8, with unchanged D1 counts. It is still retrospective.
No prospective false-alarm rate, anytime guarantee or branch-contract
breach detector is validated.

\begin{table*}[t]
\centering\footnotesize\begin{tabular}{p{5.2cm}p{5.4cm}p{5.0cm}}
\toprule
Historical criterion / later sensitivity & Evidence & Status \\
\midrule
T1 historical-gap temperature fit & gate/choice 1.3--1.5\% vs goal 1\%; corn varies by coupling & ADOPTED; absolute calibration goal not universally met \\
T2a LCD vs B3a ($\ge$30\% hot) & corn +6.5/+2.0/--0.9/--5.8\% by $\gamma$ (10 seeds); heads $\pm$0--2\% & FAILED magnitude criterion; descriptive direction is mixed \\
T2a' LCD vs B3b (blind) & corn +40--77\%; heads 5--17\% & DESCRIPTIVE comparison with damaged B3b, not general superiority \\
T2b TOST $\gamma{=}0$ ($\pm$0.3 pts) & small gate/choice differences; corn $\bar d{=}{+}0.02$ pts, $4/10$ seeds favorable (10 seeds) & PENDING test/interval reconciliation; differences alone are not TOST \\
T3 router ($\le$1/3 B4, viol parity) & 12-cell sweep, seeds 0--2: 4--6 vs 200--2044 switches; sealed gamma 0.5 outside band & DESCRIPTIVE test sweep; no 40-cell validation of selected replacement. Nominal B4 switch cost 0.057--0.211\% of horizon \\
T4 predictive transient bound & corn drain medians 3.6--5.1\,pts overall; 5.5--7.8 on the teacher-served subset (163 windows, 40 cells) & PENDING independent RHS validation; 200 event-block draws descriptive \\
T5 capacity precondition ($\rho$) & anchored pre-campaign, no re-fitting & DECLARED capacity margin; not a percentile SLA certificate \\
Literal p99 $\le$1.5 ms SLA & student p99 1.6--1.7 ms; Schmitt 1.6--13 ms & FAILED in reported summaries \\
Retrospective H4 repair sensitivity & 39/40 corn improvements; same-net own-map baseline & DESCRIPTIVE; delayed-label availability pending; no mode advantage detected after retrospective Holm \\
H5 same-checkpoint policy sensitivity & student-only lower corn H in 27/40 (8/8/6/5 of 10 by gamma) & DESCRIPTIVE total policy comparison, not full causal mediation \\
Retrospective CUSUM sensitivity & h=.3: 30/40 detected, 5 in-sample false alarms; LOO gives 8 & PENDING prospective test; 640 s nominal, 390 s monitored spans \\
\bottomrule
\end{tabular}

\caption{Historical criteria and retrospective extensions, distinguished
by evidence base. Failed, descriptive and pending statuses do not become
passes by relabeling a metric or inspecting a test sweep. H4/H5/monitor
sensitivities are not claimed as pre-registered success criteria.}
\label{tab:scorecard}
\end{table*}

\paragraph{Limitations and remaining failures.}
No independently frozen, nonvacuous branch-aware budget has been tested
on student-selected support. The common teacher-trajectory audit cannot
fill that gap. $H$ has implementation-route and smoothing biases and
does not inherit SmoothECE guarantees. Historical bootstrap coverage,
equivalence of every stratum and sequential alarm calibration are
unvalidated. Accurate argmax reproduction is not probability identity.
Retrospective repair uses unavailable-at-split labels unless availability
is separately established; final pending labels and per-batch costs are
unavailable in saved summaries. Service-start timestamps and full served
choice/corn probabilities are not recorded. The literal SLA fails even
for student-only. $\gamma=1$ degenerates gate, and empty deployed hot
strata are unsupported comparisons, not successful zero-error cells.

External evidence remains synthetic, single-platform and threshold-router
based. Ten training seeds do not substitute for independent arrival/service
traces. Learned routers, cross-platform workloads and live traffic have
not been evaluated under matched resource/label budgets. Many fewer
nearly-free nominal switches is not net economic benefit. The original
teacher has not actually been served in the adapted-checkpoint fallback.
The weighted slice diagnostic is not full causal mediation.

\paragraph{Roadmap with frozen tests.}
First record actual service-start/batch costs and full probabilities, and
evaluate $T_0/W/S$ on the same complete selected traces, retaining eligible
counts, empty cells and checkpoint/output-map versions. Recompute the
global fixed-trajectory replacement before studying trajectory effects.
Use prior data only to design separate $W/S$ budgets and support regions;
freeze these, policies, utility margins, label budgets and primary
comparisons before \emph{new} arrival/service traces. Fit recalibrators only
from $t_{\mathrm{dep}}+\tau_L$-available labels, including an optimized
intercept-only baseline, and test mode value against mixture with the same
labels. Separate monitor fitting, threshold selection and prospective
testing over longer horizons. Require useful probability/ordinal utility,
net measured costs and a satisfied percentile SLA against same-student
own-map baselines. Report failure or vacuity if those criteria fail;
do not reselect a regime or weaken baselines after seeing test outcomes.

\section{Conclusion}
CA-MHD provides a reproducible diagnostic setting for coupled-load
inference and exposes important branch-identity, metric, support and
latency pitfalls. The historical common-trajectory deficit is small,
but a deployed calibration contract is not thereby certified. Routing
shows no demonstrated net advantage under a satisfied SLA in this study;
comparators and per-cell directions differ, and retrospective repair and
monitoring are not prospective guarantees. The proved residual
composition theorem identifies sufficient conditions to test next, not
conditions the current checkpoints already meet. A nonvacuous independent
budget and a regime with material, matched-cost utility remain open.

\bibliographystyle{ACM-Reference-Format}
\bibliography{refs}

@inproceedings{clipper2017,
  author    = {Crankshaw, Daniel and Wang, Xin and Zhou, Giulio and Franklin, Michael J. and Gonzalez, Joseph E. and Stoica, Ion},
  title     = {Clipper: A Low-Latency Online Prediction Serving System},
  booktitle = {14th USENIX Symposium on Networked Systems Design and Implementation (NSDI)},
  year      = {2017}
}

@inproceedings{infaas2021,
  author    = {Romero, Francisco and Li, Qian and Yadwadkar, Neeraja J. and Kozyrakis, Christos},
  title     = {{INFaaS}: Automated Model-less Inference Serving},
  booktitle = {2021 USENIX Annual Technical Conference (USENIX ATC)},
  year      = {2021}
}

@inproceedings{laskaridis2020spinn,
  title   = {{SPINN}: Synergistic Progressive Inference of Neural Networks over Device and Cloud},
  author  = {Laskaridis, Stefanos and Venieris, Stylianos I. and Almeida, Mario and Leontiadis, Ilias and Lane, Nicholas D.},
  booktitle = {International Conference on Mobile Computing and Networking (MobiCom)},
  year    = {2020},
  doi     = {10.1145/3372224.3419194},
  note    = {arXiv:2008.06402}
}

@inproceedings{yu2019slimmable,
  title     = {Slimmable Networks},
  author    = {Yu, Jiahui and Yang, Linjie and Xu, Ning and Yang, Jianchao and Huang, Thomas},
  booktitle = {International Conference on Learning Representations (ICLR)},
  year      = {2019}
}

@inproceedings{yu2019universally,
  title     = {Universally Slimmable Networks and Improved Training Techniques},
  author    = {Yu, Jiahui and Huang, Thomas S.},
  booktitle = {IEEE/CVF International Conference on Computer Vision (ICCV)},
  pages     = {1803--1811},
  year      = {2019}
}

@inproceedings{cai2020ofa,
  title   = {Once-for-All: Train One Network and Specialize It for Efficient Deployment},
  author  = {Cai, Han and Gan, Chuang and Wang, Ji and Han, Song},
  booktitle = {International Conference on Learning Representations (ICLR)},
  year    = {2020}
}

@inproceedings{devvrit2024matformer,
  title     = {MatFormer: Nested Transformer for Elastic Inference},
  author    = {Devvrit and Kudugunta, Sneha and Kusupati, Aditya and Dettmers, Tim and Chen, Kaifeng and Dhillon, Inderjit and Tsvetkov, Yulia and Hajishirzi, Hannaneh and Kakade, Sham and Farhadi, Ali and Jain, Prateek},
  booktitle = {Advances in Neural Information Processing Systems (NeurIPS)},
  year      = {2024},
  note      = {arXiv:2310.07707}
}

@inproceedings{gujrati2020clockwork,
  author    = {Gujarati, Arpan and Karimi, Reza and Alzayat, Safya and Hao, Wei and Kaufmann, Antoine and Vigfusson, Ymir and Mace, Jonathan},
  title     = {Serving {DNNs} like Clockwork: Performance Predictability from the Bottom Up},
  booktitle = {14th USENIX Symposium on Operating Systems Design and Implementation (OSDI)},
  pages     = {443--462},
  year      = {2020}
}

@article{chen2023frugalgpt,
  title   = {FrugalGPT: How to Use Large Language Models While Reducing Cost and Improving Performance},
  author  = {Chen, Lingjiao and Zaharia, Matei and Zou, James},
  journal = {arXiv preprint arXiv:2305.05176},
  year    = {2023}
}

@inproceedings{ovadia2019trust,
  title     = {Can You Trust Your Model's Uncertainty? Evaluating Predictive Uncertainty under Dataset Shift},
  author    = {Ovadia, Yaniv and Fertig, Emily and Ren, Jie and Nado, Zachary and Sculley, D. and Nowozin, Sebastian and Dillon, Joshua V. and Lakshminarayanan, Balaji and Snoek, Jasper},
  booktitle = {Advances in Neural Information Processing Systems (NeurIPS)},
  year      = {2019}
}

@inproceedings{blasiok2023smoothece,
  title   = {Smooth {ECE}: Principled Reliability Diagrams via Kernel Smoothing},
  author  = {B{\l}asiok, Jaros{\l}aw and Nakkiran, Preetum},
  booktitle = {International Conference on Learning Representations (ICLR)},
  year    = {2024},
  note    = {arXiv:2309.12236}
}

@inproceedings{blasiok2023unifying,
  title     = {A Unifying Theory of Distance from Calibration},
  author    = {B{\l}asiok, Jaros{\l}aw and Gopalan, Parikshit and Hu, Lunjia and Nakkiran, Preetum},
  booktitle = {Annual ACM Symposium on Theory of Computing (STOC)},
  year      = {2023},
  note      = {arXiv:2211.16886}
}

@inproceedings{ross2011dagger,
  title   = {A Reduction of Imitation Learning and Structured Prediction to No-Regret Online Learning},
  author  = {Ross, St{\'e}phane and Gordon, Geoffrey and Bagnell, Drew},
  booktitle = {AISTATS},
  year    = {2011}
}

@inproceedings{guo2017calibration,
  author    = {Guo, Chuan and Pleiss, Geoff and Sun, Yu and Weinberger, Kilian Q.},
  title     = {On Calibration of Modern Neural Networks},
  booktitle = {International Conference on Machine Learning (ICML)},
  year      = {2017}
}

@article{hinton2015distilling,
  author  = {Hinton, Geoffrey and Vinyals, Oriol and Dean, Jeff},
  title   = {Distilling the Knowledge in a Neural Network},
  journal = {arXiv preprint arXiv:1503.02531},
  year    = {2015}
}

@inproceedings{kendall2018multitask,
  author    = {Kendall, Alex and Gal, Yarin and Cipolla, Roberto},
  title     = {Multi-Task Learning Using Uncertainty to Weigh Losses for Scene Geometry and Semantics},
  booktitle = {IEEE Conference on Computer Vision and Pattern Recognition (CVPR)},
  year      = {2018}
}

@article{shi2023corn,
  author  = {Shi, Xintong and Cao, Wenzhi and Raschka, Sebastian},
  title   = {Deep Neural Networks for Rank-Consistent Ordinal Regression Based on Conditional Probabilities},
  journal = {Pattern Analysis and Applications},
  volume  = {26},
  number  = {3},
  pages   = {941--955},
  year    = {2023},
  doi     = {10.1007/s10044-023-01181-9}
}

@article{chow1970,
  author  = {Chow, C. K.},
  title   = {On Optimum Recognition Error and Reject Tradeoff},
  journal = {IEEE Transactions on Information Theory},
  volume  = {16},
  number  = {1},
  pages   = {41--46},
  year    = {1970}
}

@inproceedings{geifman2017selective,
  author    = {Geifman, Yonatan and El-Yaniv, Ran},
  title     = {Selective Classification for Deep Neural Networks},
  booktitle = {Advances in Neural Information Processing Systems (NeurIPS)},
  year      = {2017}
}

@inproceedings{teerapittayanon2016branchynet,
  author    = {Teerapittayanon, Surat and McDanel, Bradley and Kung, H. T.},
  title     = {BranchyNet: Fast Inference via Early Exiting from Deep Neural Networks},
  booktitle = {International Conference on Pattern Recognition (ICPR)},
  year      = {2016}
}

@inproceedings{schuster2022confident,
  author    = {Schuster, Tal and Fisch, Adam and Gupta, Jai and Dehghani, Mostafa and Bahri, Dara and Tran, Vinh Q. and Tay, Yi and Metzler, Donald},
  title     = {Confident Adaptive Language Modeling},
  booktitle = {Advances in Neural Information Processing Systems (NeurIPS)},
  year      = {2022}
}

@inproceedings{gibbs2021adaptive,
  author    = {Gibbs, Isaac and Cand{\`e}s, Emmanuel},
  title     = {Adaptive Conformal Inference Under Distribution Shift},
  booktitle = {Advances in Neural Information Processing Systems (NeurIPS)},
  year      = {2021}
}

@inproceedings{ding2024hybrid,
  author    = {Ding, Dujian and Mallick, Ankur and Wang, Chi and Sim, Robert and Mukherjee, Subhabrata and R{\"u}hle, Victor and Lakshmanan, Laks V. S. and Awadallah, Ahmed H.},
  title     = {Hybrid LLM: Cost-Efficient and Quality-Aware Query Routing},
  booktitle = {International Conference on Learning Representations (ICLR)},
  year      = {2024}
}

@article{ong2024routellm,
  author  = {Ong, Isaac and Almahairi, Amjad and Wu, Vincent and Chiang, Wei-Lin and Wu, Tianhao and Gonzalez, Joseph E. and Kadous, M. Waleed and Stoica, Ion},
  title   = {RouteLLM: Learning to Route LLMs with Preference Data},
  journal = {arXiv preprint arXiv:2406.18665},
  year    = {2024}
}

@article{jain2024router,
  author = {Jain, Kunal and Parayil, Anjaly and Mallick, Ankur and Choukse, Esha and Qin, Xiaoting and Zhang, Jue and Goiri, {\'I}{\~n}igo and Wang, Rujia and Bansal, Chetan and R{\"u}hle, Victor and Kulkarni, Anoop and Kofsky, Steve and Rajmohan, Saravan},
  title = {Intelligent Router for {LLM} Workloads: Improving Performance Through Workload-Aware Load Balancing},
  journal = {arXiv preprint arXiv:2408.13510},
  year = {2024},
  note = {Version 2, January 2025},
  url = {https://arxiv.org/abs/2408.13510}
}

@article{zhang2025nexussched,
  author = {Zhang, Yue and Chen, Yuansheng and Mo, Xuan and Xi, Alex and Li, Jialun and Wu, WeiGang},
  title = {A Predictive and Synergistic Two-Layer Scheduling Framework for {LLM} Serving},
  journal = {arXiv preprint arXiv:2509.23384},
  year = {2025},
  url = {https://arxiv.org/abs/2509.23384}
}

@article{farran2026monitor,
  author = {Farran, Tristan},
  title = {When Your Model Stops Working: Anytime-Valid Calibration Monitoring},
  journal = {arXiv preprint arXiv:2603.13156},
  year = {2026},
  url = {https://arxiv.org/abs/2603.13156}
}

@article{tumkur2026route,
  author = {Tumkur, Srikanta Datta and Iyer, Jay and Simhadri, Mehar and Kumar, Sai Pavan and Kumar, Sai Kapil and Nampelly, Ramesh},
  title = {Calibrate, Then Route: A Measured Study of Learned Request Routing for Disaggregated {LLM} Serving},
  journal = {arXiv preprint arXiv:2609.16206},
  year = {2026},
  url = {https://arxiv.org/abs/2609.16206}
}
\appendix
\section{Proofs and Exact Accounting}
\label{app:theory}
\paragraph{Proof of Proposition~\ref{prop:mix}.}
For a nonnegative kernel with $\int_0^1K(t,c)dt=1$, write
$g_P(t)=\mathbb E_P K(t,C)(Y-C)$.
Linearity in the law and the triangle inequality give
\[
 \left\|g_{\sum_u w_uP_u}\right\|_1
 =\left\|\sum_u w_ug_{P_u}\right\|_1
 \le\sum_u w_u\|g_{P_u}\|_1.
\]
Tonelli and unit mass imply
$\|g_P\|_1\le\mathbb E|Y-C|\le1$.
The reflected Gaussian in Eq.~\ref{eq:residual} has unit mass because it
is the density of a Gaussian folded onto $[0,1]$ by reflection at
consecutive integers. For the served law, condition on the actual
selected cells within eligibility $V_h$. Its component weights are
$w_u=\Pr(U=u\mid V_h)$, and its components are the laws of
$(C_{B(u)},Y_{B(u)})$ on those cells. Apply the inequality and
$R(B(u)\mid u,V_h)\le a_u+b_{B(u),h,u}$ to obtain the stated bound.
This step requires the teacher and branch bounds on the \emph{same}
selected law, not a marginal or teacher-only trajectory.

If $\mathbb E[Y_{B(u)}\mid C_{B(u)},U=u,V_h]=C_{B(u)}$, then
conditioning on confidence makes each residual field zero. The tower
property also gives $\mathbb E[Y_{\mathrm{served}}\mid
C_{\mathrm{served}},V_h]=C_{\mathrm{served}}$ after mixing.
To show marginal calibration insufficient, take four equiprobable
feature values with true label one. Predictor A has confidence 0.75 and
is wrong only at the first feature; B also has confidence 0.75 and is
wrong only at the second. Each is marginally calibrated. Select A at
features one/three and B at two/four. The served confidence is 0.75,
accuracy 0.5, and residual error 0.25 by kernel normalization. Selection
does not consult labels. This proves both the sufficient and negative
parts of the proposition.\hfill$\square$

\paragraph{Why the theorem does not certify historical $H$.}
For equal mass at confidences 0.5 and 0.6 with respective conditional
correctness 0.5 and 0.6, each component's pairwise $H$ is zero.
At $\kappa=0.05,r=5$, cross-atom weight is $a=e^{-2}$. The fits are
$(0.5+0.6a)/(1+a)$ and $(0.6+0.5a)/(1+a)$, so the mixture gap is
$0.1a/(1+a)=0.0119202922>0$. Neither zero-on-calibrated-laws nor
convex mixing holds for this historical definition. The smoothing bias
is explicit in
\[
 c-\operatorname{NW}(Y)(c)
 =-\operatorname{NW}(Y-C)(c)+c-\operatorname{NW}(C)(c).
\]
Changing the outer integration measure is another difference from $R$.
For a fixed empirical sample with self-inclusion, $\kappa\downarrow0$
retains only confidence ties, giving
$\sum_v p_v|v-\mathbb E[Y\mid C=v]|$; with distinct confidences this
is $n^{-1}\sum_i|c_i-y_i|$. Indeed, for sufficiently small scale every
other distinct atom lies outside the truncated radius, while tied weights
are one. As $\kappa\uparrow\infty$, all weights tend to one, yielding
$\mathbb E|C-\mathbb E Y|\ge|\mathbb E C-\mathbb E Y|$ by the triangle
inequality, not equality in general. Finite summation justifies these
limits for the pairwise route only. They are not population consistency
results or limits of the distinct large-pool grid route.

\paragraph{Proof of Proposition~\ref{prop:deadband}.}
For any $a>0$,
$(a c_{\mathrm{sw}})/(a\Delta\tau_{\mathrm{ev}})
=c_{\mathrm{sw}}/\Delta\tau_{\mathrm{ev}}$.
Multiplying this unchanged ratio by the same design multiplier and
applying the same dimensionless normalization/clipping map leaves its
output unchanged. A floor on a dimensional time difference need not
scale: $\max(a\Delta\tau_{\mathrm{ev}},f)$ need not equal
$a\max(\Delta\tau_{\mathrm{ev}},f)$. Therefore the claim about actual
thresholds requires that such floors be inactive (or co-scaled), and
does not establish the un-clipped width when a clip is active.
Dwell, arrival timing and the SLA are not in this algebra and receive
no invariance claim.\hfill$\square$

\paragraph{Proof of Proposition~\ref{prop:switch}.}
Order switch times $t_1<\cdots<t_N$. Every consecutive pair requires
the signal to traverse from one threshold to the other, hence at least
$d_q$ variation on $[t_i,t_{i+1}]$. The intervals have disjoint interiors,
so $(N-1)d_q\le V_q$. Dwell requires
$t_{i+1}-t_i\ge\tau_d$, giving $(N-1)\tau_d\le H$.
Both bounds also hold for zero or one switch. Overrides/resets that
bypass crossings are excluded, and the initial constant cannot be
omitted.\hfill$\square$

\paragraph{Switch-cost accounting.}
Total switch time is $\sum_j c_{\mathrm{sw},j}$.
For $N$ events with constant student cost $\tau_S$ and constant switch
cost, switch-only overhead relative to student service is
$(N_{\mathrm{sw}}/N)c_{\mathrm{sw}}/\tau_S$, \emph{not} division by
the wide--narrow difference. On a fixed batch sequence the reference is
$\sum_j\tau_S(b_j)$; total extra time also includes
$\sum_j[\tau_{\mathrm{mode}}(b_j)-\tau_S(b_j)]$. Different policies
require their actual batch sequences. Horizon, busy-service time and
end-to-end latency are distinct denominators.

\paragraph{Proof of Proposition~\ref{prop:fluid}.}
Set $a(t)=\lambda(t)-\mu(t)$ and
$x(t)=Q(0)+\int_0^t a(s)ds$. The minimal reflection regulator is
$-\min(0,\inf_{0\le u\le t}x(u))$. Equivalently,
\[
 Q(t)=\max\left\{Q(0)+\int_0^t a(s)ds,
              \sup_{0\le u\le t}\int_u^t a(s)ds\right\}.
\]
Let $\sigma=\int_0^T(\mu-\lambda)dt>0$. Every full-period integral of
$a$ is $-\sigma$. For $t\ge T$, deleting full periods from the start
of an interval in the supremum increases its integral. Thus the
supremum equals
$\sup_{t-T\le u\le t}\int_u^t a(s)ds$.
This quantity is nonnegative (choose $u=t$), periodic and bounded by
$\int_0^T\lambda(s)ds$. The initial-backlog term loses $\sigma$ each
period; its within-period fluctuation is bounded by
$\int_0^T|a(s)|ds$. It eventually falls below zero uniformly in phase,
leaving exactly the bounded periodic supremum. If the integral of net
capacity is negative instead, the initial-backlog term grows linearly
along period endpoints, forcing unbounded backlog. If $a=0$ identically,
any finite constant backlog is possible, demonstrating that equality
does not imply instability.\hfill$\square$

\paragraph{No percentile implication.}
With arrivals spaced 3\,ms apart and a periodic sequence containing
98\% services of 0.1\,ms and 2\% of 2\,ms, there is no waiting,
mean service is 0.138\,ms and p99 is 2\,ms. Stability and a mean below
1.5\,ms therefore do not imply the target percentile. For rounded
full-batch teacher capacity 32,300/s, the stated arrival cycle has mean
12,500/s yet a burst backlog of $(35,000-32,300)\times2=5,400$ events.
It drains in $5,400/(32,300-5,000)\approx0.198$\,s in this illustrative
fluid model. Full-batch speed is not a uniform bound for the random-batch DES.

\paragraph{Exact counterfactual accounting, not slice additivity.}
For any functional $F$ recomputed on the \emph{whole} population, a
Schmitt trajectory $\tau_\pi$, a same-checkpoint student-only trajectory
$\tau_0$, and the served predictor rule $\Pi$, adding and subtracting
$F(S;\tau_\pi)$ yields the exact identity
\begin{align*}
 F(\Pi;\tau_\pi)-F(S;\tau_0)
 ={}&[F(\Pi;\tau_\pi)-F(S;\tau_\pi)]\\
    &+[F(S;\tau_\pi)-F(S;\tau_0)].
\end{align*}
The first term replaces predictions in fixed states; the second changes
the trajectory for S. This accounting convention is not unique causal
mediation; reversing the path can assign interactions differently.
For a direct nonadditivity example, let all ten confidences be 0.6,
five wide-slice observations correct and five other observations wrong.
Both $H$ and $R$ are 0.1 globally. Replacing the five correct wide
predictions by five wrong student predictions gives global error 0.6:
served-minus-counterfactual is $-0.5$, but the weighted slice difference
is only $\tfrac12(0.4-0.6)=-0.1$. A product of slice gaps is therefore
not a global replacement effect, even for $R$.

\section{Historical Design and Reproduction}
\label{app:prereg}
The original pre-registration used seeds 0--2, four coupling levels,
bandwidth 0.05 and $\pm2$\,s windows with minimum 200 events. It proposed
predictive transient checks and equivalence/multiplicity procedures that
are not satisfied merely by the current descriptive aggregates.
The ten-seed extension, H4/H5 and monitoring sensitivities are subsequent
analyses, not retroactive preregistration. Recorded pilot amendments:
A1 rescaled service 2.5$\to$1 and arrivals by 2.5; A2 allowed
historical-gap temperature fitting after test-informed investigation;
A3 replaced live service timing by a hardware-table lognormal clock.
The same ratio scaling preserved nominal loads, not every future SLA
or numerical floor/clipping behavior. A3 measures service-table CVs but
models switch variability and excludes staging/repair runtime.

Paper tables read historical JSON through
\code{scripts/tables_paper.py}, \code{scripts/tables_contract_f2.py}
and \code{scripts/tables_frontier.py}; figures use
\code{scripts/make_report.py} and \code{scripts/fig_transient.py}.
Legacy identifiers are retained for traceability. Recomputing plots/tables
does not authorize retraining or replacing historical JSON.
The integrated proofs are in \code{paper/theory_verified.tex}; any future
source bundle must include this input. No internal audit reports or
standalone working theory document are release dependencies.

\paragraph{Artifact pointers.}
Historical project artifacts are indexed at
\code{https://github.com/vectrayx/ca-mhd}; weights at
\code{https://huggingface.co/jsantillana/ca-mhd-weights}; dataset snapshot
DOI 10.5281/zenodo.23540789. These historical snapshots need not contain
this local revision.

\section{Prospective FIFO Mechanism and Filtration Checks}
\label{app:m3}
This appendix records a prospective synthetic mechanism separately from the
CA--MHD historical results.  It is not evidence that the deployed system
satisfies a calibration contract, an SLA, or an active-audit guarantee.

The mechanism has an observed finite sequence of $A,B,\mathrm{recovery}$
blocks and a single FIFO audit server.  Auditing $A$ occupies two service
units and prevents the contemporaneous $B$ event from being audited; auditing
$B$ occupies one unit.  A label is requested before its value is generated
and becomes available only a fixed number of slots after audit work completes.
The loss-unit target is kept separate from work: for fixed weights $w_z$,
\[
 U=\sum_{z\in\{A,B\}}w_z\mu_z-\lambda,
 \qquad \mu_z=\mathbb{E}[L_S-L_T\mid z].
\]
Thus computational work is not silently subtracted from a bounded utility.

For each cell, a policy must use only the filtration generated by public
queue state, completed work, prior mature labels, and fresh pre-label
randomization.  A conservative future certificate would maintain separate
confidence sequences by origin cell and combine their lower bounds; it cannot
filter labels by the queue at their later maturity time.  A proof that these
sequences remain valid for an adaptive delayed policy is still required.

As an implementation check, the FIFO enumerator verifies that an $A$ audit
blocks its following $B$ audit, that a separate audit server removes precisely
that blockage, and that work budgets and delay accounting are respected.  In a
fixed 200-replication synthetic smoke, uniform querying produced on average
25.13 mature $B$ labels, while a predeclared $B$-prioritizing allocation
produced 74.70 at essentially matched work expenditure.  With labels added
only after this physical check, the corresponding synthetic positive-sign
smoke produced positive certificates in 0.025 versus 0.730 of replications;
both policies produced none in the negative-sign smoke.  These are debugging
and feasibility observations, not coverage, power, utility, or novelty
claims.  In particular, they do not repair the historical branch, label-delay,
or SLA limitations of CA--MHD.

\end{document}